\documentclass[5p,times,authoryear]{elsarticle}

\usepackage{epsfig}
\usepackage{graphicx}

\usepackage{amssymb}
\usepackage{amsmath}

\setcitestyle{square,numbers}

\usepackage{xcolor}
\usepackage{amsthm}
\usepackage{mathrsfs} 
\usepackage{tabularx}
\usepackage{float}
\usepackage{times}

\usepackage{adjustbox} 

\def\TO{\mathcal{T}_I}
\def\R{\mathbb{R}}
\def\U{\mathbb{U}}
\def\V{\mathbb{V}}

\def\GO{\mathcal{G}_I}
\def\G{\mathcal{G}}

\def\RR{\mathscr{R}}

\def\T{\mathcal{T}}
\newcommand{\Sk}{\hat{\mathbb{S}}^{(k)}}
\newcommand{\Vk}{\hat{\mathbb{V}}^{(k)}}

\newcommand{\bigo}{\mathcal{O}}

\begin{document}

\begin{frontmatter}

\title{Invariance encoding in sliced-Wasserstein space for image classification with limited training data}

\author[affil_lab,affil_bme]{Mohammad Shifat-E-Rabbi\corref{cor1}} 
\ead{mr2kz@virginia.edu}

\author[affil_lab,affil_ece]{Yan Zhuang}
\ead{yz8bk@virginia.edu}

\author[affil_lab,affil_bme]{Shiying Li}
\ead{sl8jx@virginia.edu}

\author[affil_lab,affil_ece]{Abu Hasnat Mohammad Rubaiyat}
\ead{ar3fx@virginia.edu}

\author[affil_lab,affil_ece]{Xuwang Yin}
\ead{xy4cm@virginia.edu}

\author[affil_lab,affil_bme,affil_ece]{Gustavo K. Rohde}
\ead{gustavo@virginia.edu}

\cortext[cor1]{Corresponding author.}

\address[affil_lab]{Imaging and Data Science Laboratory, University of Virginia, Charlottesville, VA 22908, USA}
\address[affil_bme]{Department of Biomedical Engineering, University of Virginia, Charlottesville, VA 22908, USA}
\address[affil_ece]{Department of Electrical and Computer Engineering, University of Virginia, Charlottesville, VA 22908, USA}

\begin{abstract}
Deep convolutional neural networks (CNNs) are broadly considered to be state-of-the-art generic end-to-end image classification systems. However, they are known to underperform when training data are limited and thus require data augmentation strategies that render the method computationally expensive and not always effective. Rather than using a data augmentation strategy to encode invariances as typically done in machine learning, here we propose to mathematically augment a nearest subspace classification model in sliced-Wasserstein space by exploiting certain mathematical properties of the Radon Cumulative Distribution Transform (R-CDT), a recently introduced image transform. We demonstrate that for a particular type of learning problem, our mathematical solution has advantages over data augmentation with deep CNNs in terms of classification accuracy and computational complexity, and is particularly effective under a limited training data setting. The method is simple, effective, computationally efficient, non-iterative, and requires no parameters to be tuned. Python code implementing our method is available at \cite{software}. Our method is integrated as a part of the software package PyTransKit, which is available at \cite{package}.
\end{abstract}


\end{frontmatter}



\section{Introduction}
\begin{figure}[t]
\begin{center}
\includegraphics[width=0.65\linewidth]{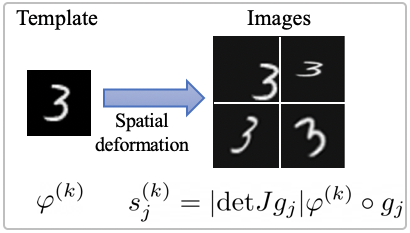}
\end{center}
  \caption{In many classification problems, images in a class can be thought of being an instance of a template observed under a set of spatial deformations.}
\label{fig:f00}
\end{figure}

Image classification methods occupy a predominant place in data sciences, given their inherent link to numerous machine learning and computer vision applications \cite{basu2014detecting,kundu2018discovery,mikolajczyk2018data,o2019comparing,hussain2017differential,wang2017effectiveness,wong2016understanding,simard2003best,lecun1998gradient,murphy2000towards,bloice2019biomedical,azulay2018deep,kundu2020enabling,rabbi2017speckle,nishikawa2021massive,shifat2020cell}. In many image classification problems, image classes can be thought of being an instance of a template observed under a set of spatial deformations. For example, consider the classes of the MNIST dataset \cite{lecun1998gradient}. Each image in a class can be thought of being an image of a prototype digit with a transformation applied to it (such as translation, scaling, shear, rotation, higher-order deformations, and others; see Fig.~\ref{fig:f00}). Other examples of this category of image classification problems include detecting the protein localization patterns within a cell \cite{murphy2000towards}, classifying the nuclear structures from the fluorescence measurements of a population of cells \cite{basu2014detecting}, and profiling the distribution of gray
matter within a brain as depicted through MRI \cite{kundu2018discovery}, among others.

\begin{figure*}[!hbt]
\begin{center}
\includegraphics[width=1.0\linewidth]{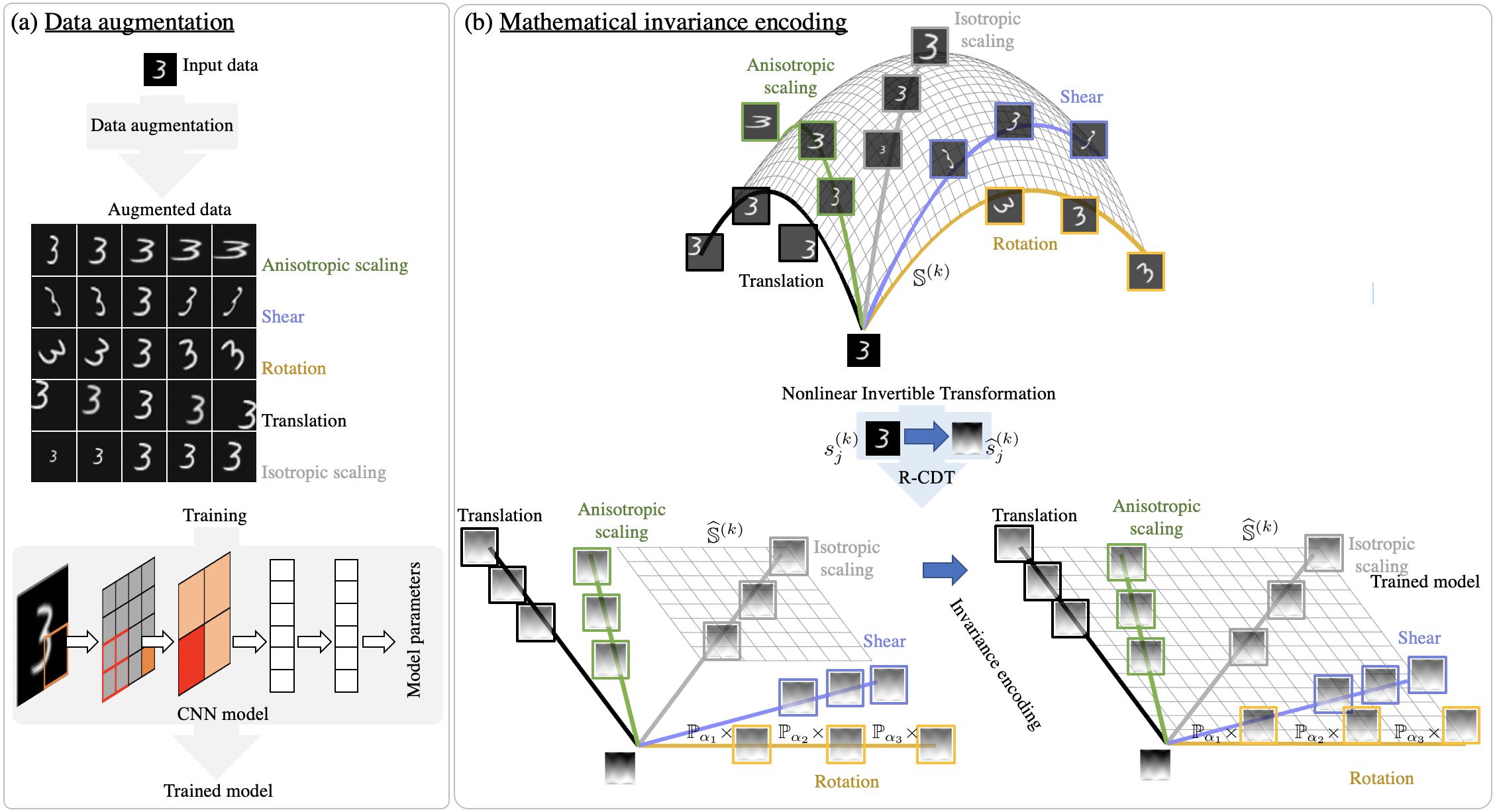}

  \caption{System diagrams outlining the data augmentation-based methods and the proposed method. (a) Data augmentation-based methods augment the training set by artificially applying known transformations to the original training set. (b) The proposed invariance encoding method models the underlying data space (represented by grey grid lines) corresponding to known transformations to learn invariances to those transformations. The R-CDT renders data space convex and enables it to be modeled with a linear subspace. The invariance encoding framework expands the subspace to incorporate invariances to desired transformations.}
  \end{center}
\label{fig:f01}
\end{figure*}

Over the past few decades, image classification methods have evolved from feature engineering-based methods relying on hand-tailored numerical features \cite{prewitt1966analysis,fisher1936use,boland2001neural,orlov2008wnd,ponomarev2014ana,nosaka2014hep} to hierarchical (deep) convolutional neural network-based (CNN) methods utilizing a series of computational layers \cite{simonyan2014very,he2016deep,szegedy2015going,lecun2015deep,wang2017effectiveness,wong2016understanding,simard2003best,lecun1998gradient,deng2014deep}. CNNs have recently emerged as leading classification methods for several reasons \cite{simonyan2014very,he2016deep,szegedy2015going,lecun2015deep,deng2014deep,szegedy2016rethinking,krizhevsky2012imagenet}. They provide a framework for end-to-end learning bypassing the feature engineering process, often decreasing the time and expenses related to bringing classification systems into production \cite{lecun1998gradient,lecun2015deep}. They obtain high accuracy in several image classification tasks \cite{simonyan2014very,he2016deep,szegedy2015going} and offer feasibility to be implemented in parallel utilizing graphical processing units (GPU) \cite{strigl2010performance,gu2018recent,potluri2011cnn}, among other improvements and conveniences over feature engineering methods. Recent improvements in computing power \cite{wu2019high}, the availability of annotated data \cite{gu2018recent} and open-source software \cite{gulli2017deep} have also contributed to the usability of CNNs. However, it is broadly understood that CNN-based methods require large amounts of data for training \cite{liu2018unsupervised,jang2020etri,padarian2019using}, are computationally expensive \cite{bappy2016cnn,brosch2013manifold,mousavian20173d}, time-consuming \cite{huang2016deep}, require careful parameter and hyper-parameter tuning \cite{gu2018recent,shankar2020hyperparameter,zhang2019deep,neary2018automatic}, and are often vulnerable against out-of-distribution samples and attacks \cite{shifat2021radon,rubaiyat2021nearest,lee2018simple,ren2020adversarial,akhtar2018threat}. 

As mentioned above, and in more detail in \cite{aghamaleki2019multi,fitzgerald2018training,shorten2019survey,wang2020inconsistent}, deep CNNs are known to underperform when the number of training images is limited. Building classification systems using limited training data is desirable in many applications such as medicine \cite{ching2018opportunities}, radiology \cite{wang2020inconsistent}, radar engineering \cite{wang2015application}, remote sensing \cite{pal2012evaluation}, and numerous others, given the difficulty and cost associated with acquiring high-quality labeled image data. Applications to characterize differences between the classes of organ-scale medical images (e.g., brain MRIs, lung CTs), or cellular structures (e.g., confocal microscopy), for example, are often faced with the task of decoding information based on a few hundred samples or even fewer \cite{cattell2016classification}. A common approach to overcome the shortage of training data is to augment the training set with known transformations such as affine transformations \cite{mikolajczyk2018data,o2019comparing,hussain2017differential,wang2017effectiveness,wong2016understanding,simard2003best,bloice2019biomedical}; elastic deformations \cite{wong2016understanding,simard2003best,bloice2019biomedical}; random distortions \cite{o2019comparing,hussain2017differential,wang2017effectiveness,bloice2019biomedical}; color, brightness, contrast variation \cite{hussain2017differential,bloice2019biomedical}; other geometric transformations \cite{bloice2019biomedical}; and so forth. However, even with such data augmentation procedures in place, deep CNNs may still struggle to achieve invariances to the desired image transformations \cite{azulay2018deep}. Moreover, data augmentation expands the size of the training set and thereby substantially increases the computational complexity of training the model \cite{shen2021low,taylor2018improving}.

One possible way to overcome the limitations associated with data augmentation strategies in machine learning is to incorporate knowledge of image transformations into the mathematical formulation of the classification method, with the end goal of rendering the output of the classification system invariant to known transformations, as was done in \cite{shifat2021radon}. The method proposed in \cite{shifat2021radon} utilizes a combination of a nearest subspace (NS) classifier \cite{li2013nearest,liu2011k} and the Radon Cumulative Distribution Transform (R-CDT) \cite{kolouri2016radon}, and provides the possibility to solve the classification problem depicted in Fig.~\ref{fig:f00} in a mathematically principled way. The R-CDT NS method classifies a test image by finding the nearest set to the test sample in the sliced-Wasserstein distance sense. Each set in this case, corresponds to a particular class and is defined as an instance of a template for that class observed under some random image transformations. The approach described in \cite{shifat2021radon} takes advantage of the fact that the R-CDT can render such sets convex and thus can be modeled effectively with a linear subspace, and can be learned with simple linear (e.g., PCA) methods using training data.


The R-CDT-NS method is simple, non-iterative, and for the type of problems depicted in Fig.~\ref{fig:f00} provides similar or better accuracy results compared with the state-of-the-art neural network-based methods requiring orders of magnitude fewer floating-point operations (FLOPS) \cite{shifat2021radon}. It is mathematically coherent, understandable, does not require hyper parameter tuning, and has been demonstrated to have clear advantages under the out-of-distribution experimental setup. One attractive property of the R-CDT-NS method \cite{shifat2021radon} (which has also been utilized for solving 1D signal classification problems \cite{rubaiyat2021nearest}) is that the class generative models become linear subspaces in R-CDT domain, and can be built from available training data, or mathematical knowledge of image transformations that are known to be present in a particular problem. However, the R-CDT-NS method as proposed in \cite{shifat2021radon} does not entirely exploit the possibilities to utilize the mathematical knowledge of image transformations, which were limited to a few simple transformations such as translation and isotropic scaling only \cite{shifat2021radon}. Part of the difficulty with prescribing spatial transformations that go beyond scaling and translation in the R-CDT domain is the fact that the mathematical formulations of these in the R-CDT domain are not known. Here we provide the necessary mathematical approximation of affine transformations in the R-CDT domain, and incorporate them in the R-CDT domain, thus greatly enhancing the performance of the method when few training samples are available, while maintaining the overall mathematical structure that allows the method to perform well when numerous training examples are available. Specifically, we highlight the following contributions in this paper:




\subsubsection*{Contributions}
\begin{itemize}
    \item This paper improves the R-CDT-NS framework \cite{shifat2021radon} and thereby proposes an improved end-to-end machine learning system for the category of classification problems outlined in Fig.~\ref{fig:f00}. We provide mathematical approximations for the affine set of spatial transformations in the Radon CDT domain, which we then use to extend the method proposed in \cite{shifat2021radon}.
    
    \item The proposed method achieves high classification accuracy compared with the original R-CDT-NS method \cite{shifat2021radon} as well as the other state-of-the-art methods, especially at the low-data regime, i.e., using a low number of training samples while maintaining high accuracy performances at the high-data regime also. The results on several datasets demonstrate the advantages of the mathematical solution with respect to the traditional data augmentation-based methods. Fig.~\ref{fig:f01} demonstrates general pictorial outlines of the data augmentation-based methods and the proposed method. 
    
    \item The proposed method is simple to implement, non-iterative, convenient to use, and computationally efficient (up to 1,000,000 times savings in the computational cost can be attained as compared with the state-of-the-art methods). We compared with several state-of-the-art end-to-end classification methods and demonstrated significant improvements in computational cost. 
    
    \item The proposed method retains high classification accuracy under challenging experimental scenarios, such as out-of-distribution setup -- meaning the method generalizes to previously unseen data samples. Our method is integrated as a part of the software package PyTransKit \cite{package}. The source code implementing our method is also available at \cite{software}.

\end{itemize}

\section{Preliminaries}
Throughout this manuscript, we consider images $s$ to be square integrable functions such that $\int_{\mathbf{\Omega}_s}|s(\mathbf{x})|^2d\mathbf{x}<\infty$, where $\mathbf{x}\in\mathbf{\Omega}_s\subseteq\mathbb{R}^2$. Images are denoted $s^{(k)}$ when class information is available, where the superscript $(k)$ denotes the class label. We also use one-to-one diffeomorphisms (one-to-one mapping functions) denoted as $g(\mathbf{x}),\mathbf{x}\in\R^2,$ for images, and $g^\theta(t),t\in\R,$ when they need to be parameterized by an angle $\theta\in[0,\pi]$. The set of all possible one-to-one diffeomorphisms from $\mathbb{R}$ to $\R$ and from $\R^2$ to $\R^2$ are denoted as $\T$ and $\TO$, respectively. Finally, we use $\circ$, $\RR(\cdot)$, and $\RR^{-1}(\cdot)$ operators to denote the composition, the forward Radon transform, and the inverse Radon transform operations, respectively.
\subsection{The Cumulative Distribution Transform (CDT)}
The cumulative distribution transform (CDT) proposed in \cite{park2018cumulative}, is an invertible nonlinear 1D signal transform from the space of smooth probability densities to the space of diffeomorphisms. The CDT morphs a positive probability density function (PDF) into another positive PDF (reference) in such a way that the Wasserstein distance between them is minimized. Formally, the CDT can be described as follows: let $s(x),x\in\Omega_s\subseteq\R$ and $r(x),x\in\Omega_r\subseteq\R$ be a given signal and a reference signal, respectively, which are appropriately normalized such that $s>0,r>0$, and $\int_{\Omega_s} s(x)dx=\int_{\Omega_r} r(x)dx=1$. The forward CDT of $s(x)$ with respect to $r(x)$ is given by the strictly increasing function $\hat{s}(x)$ that satisfies
\begin{align}
    \int_{-\infty}^{\hat{s}(x)}s(u)du=\int_{-\infty}^{x}r(u)du.\nonumber
\end{align}
Here we are using the CDT formulation used in \cite{rubaiyat2020parametric}, which slightly differs from the definition used in \cite{park2018cumulative}. The general properties of the CDT hold in both definitions. Apart from numerical issues related to interpolation, the CDT represents the input signal entirely without causing any information loss \cite{park2018cumulative}. The original signal can be reconstructed from its CDT values using the inverse formulation defined as:
\begin{align}
s(x)=\frac{d\hat{s}^{-1}(x)}{dx}r\left(\hat{s}^{-1}(x)\right), ~\mbox{and}~\hat{s}^{-1}(\hat{s}(x)) = x.\nonumber
\end{align}

\subsection{Radon Cumulative Distribution Transform (R-CDT)}
Kolouri et al. \cite{kolouri2016radon} extended the CDT framework for two-dimensional (2D) patterns through the Radon transform and proposed a new transform denoted as the Radon cumulative distribution transform (R-CDT). To compute the R-CDT of a 2D probability measure (such as an image), a family of one dimensional representations of that 2D measure is obtained through the Radon transform \cite{helgason1980radon} and then the CDT is applied over each Radon projection. Formally, let $s(\mathbf{x}),\mathbf{x}\in\Omega_s\subseteq \R^2$ and $r(\mathbf{x}),\mathbf{x}\in\Omega_r\subseteq \R^2$ denote a given image and a reference image (both appropriately normalized), respectively. The forward R-CDT of $s(\mathbf{x})$ with respect to $r(\mathbf{x})$ is given by the measure preserving function $\hat{s}(t,\theta)$ that satisfies
\begin{align}\label{eq:rcdt}
    \int_{-\infty}^{\hat{s}(t,\theta)}\tilde{s}(u,\theta)du=\int_{-\infty}^{t}\tilde{r}(u,\theta)du,~~~\forall\theta\in[0,\pi].\nonumber
\end{align}

Here, $\tilde{s}=\RR(s)$ is the Radon transform of the image $s$ defined as
\begin{eqnarray}
\tilde{s}(t,\theta)&=&\int_{\Omega_s}s(\mathbf{x})\delta(t-\mathbf{x}\cdot \mathbf{\xi}_\theta)d\mathbf{x},\nonumber
\end{eqnarray}
where $t$ is the perpendicular distance of a line from the origin, $\xi_\theta = [\cos(\theta),\sin(\theta)]^T$, and $\theta$ is the angle over which the projection is taken. Using the Fourier Slice Theorem \cite{quinto2006introduction,natterer2001mathematics}, the inverse Radon transform $s=\RR^{-1}(\tilde{s})$ can be defined as
\begin{eqnarray}
s(\mathbf{x})&=&\int_0^\pi\int_{-\infty}^{\infty}\tilde{s}(\mathbf{x}\cdot\xi_\theta-\tau,\theta)w(\tau)d\tau d\theta,\nonumber
\end{eqnarray}
where $w$ is the ramp filter (i.e.,$(\mathscr Fw) (\xi) = |\xi|, \forall \xi$ ) and $\mathscr{F}$ is the Fourier transform operator.

As in the case of the CDT, the R-CDT is also invertible and the original image can be recovered from its transform via the following inverse formula: 
\begin{align} 
s(\mathbf{x})=\RR^{-1}\left(\frac{\partial \hat{s}^{-1}(t,\theta)}{\partial t}\tilde{r}\left(\hat{s}^{-1}(t,\theta),\theta\right)\right).\nonumber
\end{align}

Note that the CDT and the R-CDT outlined above represent the input images entirely causing no information loss during the process \cite{kolouri2016radon}. However, the empirical implementation of the transforms might introduce numerical errors. We note that these numerical errors have minimal impact in solving classification problems, as shown in \cite{shifat2021radon}. The R-CDT has a couple of properties outlined below, which will be of interest when classifying images.\\\\ 
{\bf{Property 1}~{(Composition)}} Let $s(\mathbf{x})$ denotes an appropriately normalized image and let $\tilde{s}(t,\theta)$ and $\hat{s}(t,\theta)$ be the Radon transform and the R-CDT of $s(\mathbf{x})$, respectively. The R-CDT of $s_{g^\theta}=\mathscr{R}^{-1}\left(\left({g^\theta}\right)^\prime\tilde{s}\circ {g^\theta}\right)$ is given by
\begin{align}
    \hat{s}_{g^\theta}=({g^\theta})^{-1}\circ\hat{s},
\end{align}
where $\left(g^\theta\right)^\prime=dg^\theta(t)/dt$, $\tilde{s}\circ {g^\theta}:=\tilde{s}(g^\theta(t),\theta)$ and $({g^\theta})^{-1}\circ\hat{s}=({g^\theta})^{-1}(\hat{s}(t,\theta))$.
For a fixed $\theta$, $g^\theta$ can be thought of an increasing and differentiable function with respect to $t$.\\\\
{\bf{Property 2}~{(Embedding)}}
The R-CDT induces an isometric embedding between the space of images as normalized probability masures (with the sliced-Wasserstein metric ($SW_2$)) and the space of their transforms (with a weighted-Euclidean metric). In other words, for all images $s_1$ and $s_2$,
\begin{equation}\label{eq: rcdtembedding}
	SW_2^2(s_1,s_2) = \left|\left|\left(\hat{s}_1-\hat{s}_2\right)\sqrt{\tilde{r}}\right|\right|_{L_2(\Omega_{\tilde r})}^2
\end{equation}
The property above naturally connects the R-CDT and sliced-Wasserstein distances for PDFs and allows us a simple means of computing similarity among images \cite{kolouri2016radon}. We remark that throughout this manuscript, the R-CDT transforms of an image $s$ are computed with respect to a fixed reference image $r$ if a reference is not specified.

\section{Problem Statement}
We begin by noting that, in many classification problems, image classes can be thought of as an instance of a template pattern observed under some unknown spatial deformations. Consider the MNIST digit classification problem, for example. Images of each digit in a class can be thought of being a prototype template digit observed under random translations, scaling, higher-order smooth deformations, and others (see Fig.~\ref{fig:f00}). The generative model below formalizes the above statements:\\\\
{\bf{Generative model:}} 
Let $\GO\subset\TO$ be a set of smooth one-to-one transformations involving non-singular affine transformations up to a certain degree and other possible non-linear deformations. The mass (image intensity) preserving generative model for the $k$-th image class is defined to be the set
\begin{align}
\label{eq:2dgenerative_model_im}
\mathbb{S}^{(k)}=\left\{s_j^{(k)}|s_j^{(k)}=|\text{det} Jg_j|\varphi^{(k)}\circ g_j, \forall g_j\in\GO \right\},
\end{align}
where, $\varphi^{(k)}$ and $s_j^{(k)}$ denote the template and the $j$-th image, respectively, from the $k$-th image class and $\text{det} Jg_j$ denotes the determinant of the Jacobian matrix of $g_j$. In our discussion, it is useful to state an equivalent Radon-space definition of the generative model in equation~\eqref{eq:2dgenerative_model_im}. Let $\G\subset\T$ be the set of smooth deformations in the Radon space. The equivalent Radon-space generative model for the $k$-th image class is defined to be the set
\begin{align}
\label{eq:2dgenerative_model}
\mathbb{S}^{(k)}=\left\{s_j^{(k)}|s_j^{(k)}=\RR^{-1}\left(\left({g_j^\theta}\right)^\prime\tilde{\varphi}^{(k)}\circ g^\theta_j\right), \forall g^\theta_j\in\G \right\},
\end{align}
where, $\tilde{\varphi}^{(k)}$ denotes the Radon transform of the template $\varphi^{(k)}$ and $\RR^{-1}(\cdot)$ denotes the inverse Radon transform operator. With the above definition of the data generative model, we formally define the above category of image classification problem as follows:\\\\
{\bf{Classification problem:}} Let $\G\subset\T$ (or $\GO\subset\TO$) be the set of smooth deformations, and let the set of image classes $\mathbb{S}^{(k)}$ be defined as in equation~\eqref{eq:2dgenerative_model} (or equation~\eqref{eq:2dgenerative_model_im}). Given training samples $\{s^{(1)}_1, s^{(1)}_2, \cdots\}$ (class 1), $\{s^{(2)}_1, s^{(2)}_2,$ $\cdots\}$ (class 2), $\cdots$ as training data, determine the class $(k)$ of an unknown image $s$.\\

It was demonstrated in \cite{shifat2021radon} that image classes following the generative model in equation~\eqref{eq:2dgenerative_model} yield nonconvex data geometry, causing the above classification problem to be difficult to solve and necessitating nonlinear classifiers. The work in \cite{shifat2021radon} utilized the property of Radon Cumulative Distribution Transform (R-CDT) \cite{kolouri2016radon} to simplify the classification problem and provide a non-iterative solution. Here, we briefly explain the solution provided in \cite{shifat2021radon} as follows:

The solution in \cite{shifat2021radon} begins by applying the R-CDT on the generative model in equation~\eqref{eq:2dgenerative_model}. The R-CDT space generative model then becomes
\begin{eqnarray}
\Sk&=&\{\hat{s}_j^{(k)}|\hat{s}_j^{(k)}={\left(g_j^\theta\right)}^{-1}\circ \hat{\varphi}^{(k)}, \forall g_j^\theta\in\G \}. 
\end{eqnarray}
It was shown that if $\G\subset\T$ is a convex group then $\Sk$ is convex \cite{shifat2021radon}. Also, if $\mathbb{S}^{(k)} \cap \mathbb{S}^{(p)}=\varnothing$, then $\hat{\mathbb{S}}^{(k)} \cap  \hat{\mathbb{S}}^{(p)}=\varnothing$. The method in \cite{shifat2021radon} then proposes a non-iterative training algorithm by estimating subspaces $\Vk$ where $\Vk$ denotes the subspace generated by the convex set $\Sk$ as follows:
\begin{align}
    \hat{\V}^{(k)}=\mbox{span}\left(\Sk\right)=\left\{\sum_{j\in J}\alpha_j\hat{s}_j^{(k)}\mid \alpha_j\in\R, J ~\textrm {is finite}\right\}. 
\end{align}
It was also shown in \cite{shifat2021radon} that, under certain assumptions, $\Sk\cap\hat{\V}^{(p)}=\varnothing$, for $k\neq p$, i.e., the subspaces $\Vk$ do not overlap with the data class of another class. The method also explained a framework to prescribe invariance with respect to the translation operation by enhancing the subspace $\Vk$ with a spanning set corresponding to the translation deformation as follows:
\begin{align}
    \hat{\V}_A^{(k)}=\mbox{span}\left(\Sk\cup\U_T\right)
\end{align}
where $\U_T=\{u_1(t,\theta),u_2(t,\theta)\}$, with $u_1(t,\theta)=\cos\theta$ and $u_2(t,\theta)=\sin\theta$, denotes the spanning set corresponding to the translation deformation type and $\Vk_A$ denotes the enhanced subspace. Finally, the class of an unknown test sample $s$ was determined by solving
\begin{equation}
     \arg \min_k d^2(\hat{s},\hat{\V}_A^{(k)}).
    \label{eq:nearest_subspace}
\end{equation}

The method proposed in \cite{shifat2021radon} is characterized by two prime aspects of it. First, once the training examples containing a specific type of deformation are available, the method can learn other instances of that deformation type from data. Secondly, the method can encode invariances with respect to a few specific deformations in the model without explicit data augmentation when the training samples are low in number or do not contain those specific deformations. For more details, refer to \cite{shifat2021radon}.

\section{Proposed solution}
We expand upon the latter aspect of the method in \cite{shifat2021radon}. The mathematically prescribed invariances in \cite{shifat2021radon} were limited to a few simple spatial transformations such as translations and isotropic scalings (scaling by the same magnitudes in both x and y directions of the image). We expand the extent of the method and mathematically encode invariances with respect to a more complicated deformation set: the set of affine deformations (e.g., translation, both isotropic and anisotropic scaling, both horizontal and vertical shear, and rotation). The detailed explanations of the deformation types used to encode invariances and the corresponding methodologies are explained as follows:
\begin{itemize}
	\item[i)] Translation: Let $g(\mathbf{x})= \mathbf{x}-\mathbf{x_0}$ be the translation by $\mathbf{x}_0\in \R^2$ and $s_g(\mathbf{x})=|\det Jg|s\circ g=s(\mathbf{x}-\mathbf{x_0})$. Note that $Jg$ denotes the Jacobian matrix of $g$. Following \cite{kolouri2016radon} we have that $\hat s_g (t,\theta) =\hat{s}(t,\theta)+\mathbf{x}_0^T \xi_\theta$ where $\xi_\theta = [\cos(\theta),\sin(\theta)]^T$. Therefore, as in \cite{shifat2021radon}, we define the spanning set for translation as $\U_T=\{u_T^{(1)}(t,\theta),u_T^{(2)}(t,\theta)\}$, where $u_T^{(1)}(t,\theta)=\cos\theta$ and $u_T^{(2)}(t,\theta)=\sin\theta$. The spanning set $\U_T$ (and the other spanning sets described below, defined for other deformation types) is then used to enhance the subspace $\Vk$ and encode invariance to the corresponding deformation type. The methodology used to obtain the enhanced subspace $\Vk_A$ is explained in more detail in section \ref{subsec_train}.
	\item[ii)]Isotropic scaling: Let $g(\mathbf{x})=a\mathbf{x}$  and $s_g(\mathbf{x})=|Jg|s\circ g=a^2s(a\mathbf{x})$, which is the normalized dilatation of $s$ by $a$ where $a\in\R_{+}$. Then according to \cite{kolouri2016radon}, $\hat s_g (t,\theta) = \hat{s} (t,\theta)/a$, i.e. a scalar multiplication. Therefore, as in \cite{shifat2021radon}, an additional spanning set is not required here as the subspace containing $\hat{s}(t,\theta)$ naturally contains its scalar multiplication. Therefore, the spanning set for isotropic scaling is defined as $\U_D=\varnothing$.
	\item[iii)]Anisotropic scaling: Let $g(\mathbf{x})=\breve{\mathcal{D}}\mathbf{x}$ with $\breve{\mathcal{D}}=\begin{bmatrix}1/a,&0\\0,&1/b\end{bmatrix}$, $a\neq b$, and $s_g(\mathbf{x})=|Jg|s\circ g=\frac{1}{ab} s(\breve{\mathcal{D}}\mathbf{x})$, which is the normalized anisotropic dilatation of $s$ by $a$, $b$ where $a,b\in\R_{+}$. We postulate that $\hat s_g (t,\theta)$ can be approximated as $a\hat s(t,\theta)+ \alpha a\sin^2\theta\hat s(t,\theta)$, for $a<b$, and as $ b\hat s(t,\theta)+ \beta b\cos^2\theta\hat s(t,\theta)$, for $a>b$; $b/a=1+\alpha$ (for a proof see Appendix A). Therefore we define the spanning set for anisotropic scaling as $\U_{\breve{D}}=\{u_{\breve{D}}^{(1)}(t,\theta),u_{\breve{D}}^{(2)}(t,\theta)\}$, where $u_{\breve{D}}^{(1)}(t,\theta)=(\cos^2\theta)\hat{s}(t,\theta)$ and $u_{\breve{D}}^{(2)}(t,\theta)=(\sin^2\theta)\hat{s}(t,\theta)$.
	\item[iv)]Shear: Let $g_1(\mathbf{x})=\mathcal{H}_1\mathbf{x}$ with $\mathcal{H}_1=\begin{bmatrix}1,&-h\\0,&1\end{bmatrix}$, $g_2(\mathbf{x})=\mathcal{H}_2\mathbf{x}$ with $\mathcal{H}_2=\begin{bmatrix}1,&0\\-v,&1\end{bmatrix}$, and $s_{g_1}(\mathbf{x})=|Jg_1|s\circ g_1=s(\mathcal{H}_1\mathbf{x})$, $s_{g_2}(\mathbf{x})=|Jg_2|s\circ g_2=s(\mathcal{H}_2\mathbf{x})$, which are the normalized horizontal and vertical shears of $s$, respectively, by $h$ and $v$ where $h\neq v$; $h,v\in\R$. We postulate that $\hat s_{g_1} (t,\theta)$ can be approximated as $\hat s(t,\theta)+ \frac{1}{2}(h\sin(2\theta)+h^2\cos^2\theta)\hat s(t,\theta)$ and $\hat s_{g_2} (t,\theta)$ can be approximated as $\hat s(t,\theta)+ \frac{1}{2}(v\sin(2\theta)+v^2\sin^2\theta)\hat s(t,\theta)$ (for a proof see Appendix B). Therefore the spanning set for shear is defined as $\U_{H}=\{u_H^{(1)}(t,\theta),u_H^{(2)}(t,\theta)\}$, where $u_H^{(1)}(t,\theta)=(v^2\sin^2\theta+v\sin2\theta)\hat{s}(t,\theta)$ and $u_H^{(2)}(t,\theta)=(h^2\cos^2\theta+h\sin2\theta)\hat{s}(t,\theta)$.
	\item[v)]Rotation: Let $g(\mathbf{x})=\mathcal{R}\mathbf{x}$ with $\mathcal{R}=\begin{bmatrix}\cos\theta_0,&-\sin\theta_0\\\sin\theta_0,&\cos\theta_0\end{bmatrix}$, and $s_g(\mathbf{x})=|Jg|s\circ g=s(\mathcal{R}\mathbf{x})$, which is the normalized rotation of $s$ by $\theta_0$ where $\theta_0\in[0,\pi]$. Following \cite{kolouri2016radon} we have that $\hat s_g (t,\theta) =\hat{s}(t,\theta-\theta_0)$, i.e., rotation in image space results in a circular translation in angle $\theta$ in the R-CDT space, whereas our previous discussion pertains to a fixed $\theta$. Here, we encode rotation invariance in an alternate manner, in the testing phase of the method. If data contains a rotation confound and if a test image $s$ belongs to the class $(k)$ then $d^2\left(\mathbb{P}_\alpha\hat{s},\hat{\mathbb{V}}^{(k)}\right)=0~~\mbox{and}~~d^2\left(\mathbb{P}_\alpha\hat{s},\hat{\mathbb{V}}^{(l)}\right)>0;~~~k\neq l$ where, $\mathbb{P}_\alpha$ is a fixed permutation matrix that causes circular translation in $\theta$ by $\alpha\in[0,\pi]$ (which eventually causes rotation in the native image space). Therefore, the class of $s$ can be decoded by solving $\arg\min_k\min_\alpha d^2\left(\mathbb{P}_\alpha\hat{s},\hat{\mathbb{V}}^{(k)}\right)$.
\end{itemize}
\subsection{Training algorithm}
\label{subsec_train}
Using the principles laid out above, the algorithm we propose estimates the enhanced subspace $\Vk_A$ corresponding to the transform space $\Sk$ given sample data $\{s_1^{(k)},$ $s_2^{(k)},$ $\cdots\}$. Naturally, the first step is to transform the training data to obtain $\{\hat{s}_1^{(k)}, \hat{s}_2^{(k)}, \cdots \}$. We then approximate $\Vk_A$ as follows:
\begin{align}
    \label{eq:transscal} \Vk_A=\mbox{span}\left(\left\{\hat{s}_1^{(k)},\hat{s}_2^{(k)},\cdots\right\}\cup\U_A\right)
\end{align}
where $\U_A=\U_T\cup\U_D\cup\U_{\breve{D}}\cup\U_H$ denotes the combined spanning set corresponding to the translation, isotropic/anisotropic scaling, and horizontal/vertical shear deformation types; the spanning set $\U_A$ is used in equation~\eqref{eq:transscal} to obtain invariances to these deformation types. As mentioned before, the invariance to the rotation deformation is obtained alternately in the testing phase of the method (see section \ref{section_test} for details). Here, $\U_T=\{u_T^{(1)}(t,\theta),u_T^{(2)}(t,\theta)\}$, with $u_T^{(1)}(t,\theta)=\cos\theta$ and $u_T^{(2)}(t,\theta)=\sin\theta$, corresponds to the translation deformation type; $\U_D=\varnothing$ corresponds to isotropic scaling; $\U_{\breve{D}}=\{u_{\breve{D}}^{(1)}(t,\theta),u_{\breve{D}}^{(2)}(t,\theta)\}$, with $u_{\breve{D}}^{(1)}(t,\theta)=(\cos^2\theta)\left\{\hat{s}_1^{(k)}(t,\theta),\hat{s}_2^{(k)}(t,\theta),\cdots\right\}$ and $u_{\breve{D}}^{(2)}(t,\theta)=(\sin^2\theta)\left\{\hat{s}_1^{(k)}(t,\theta),\hat{s}_2^{(k)}(t,\theta),\cdots\right\}$, corresponds to anisotropic scaling; and $\U_{H}=\{u_H^{(1)}(t,\theta),u_H^{(2)}(t,\theta)\}$, with $u_H^{(1)}(t,\theta)=(v^2\sin^2\theta+v\sin2\theta)\left\{\hat{s}_1^{(k)}(t,\theta),\hat{s}_2^{(k)}(t,\theta),\cdots\right\}$ and $u_H^{(2)}(t,\theta)=(h^2\cos^2\theta+h\sin2\theta)\\\left\{\hat{s}_1^{(k)}(t,\theta),\hat{s}_2^{(k)}(t,\theta),\cdots\right\}$, corresponds to the vertical and horizontal shear deformations 
\footnote{Note that the spanning set $\Big\{(v^2\sin^2\theta+v\sin2\theta)\hat{s}\Big\}_{|v|\leq\epsilon}$ (for vertical shear, see equation~\eqref{eq:transscal}) in transform domain for some  $\epsilon$   is not linear in the sense that it is not in the $\textrm{span}\Big\{(v_0^2\sin^2\theta+v_0\sin2\theta)\hat{s}\Big\}$ for some fixed small $v_0$. The situation for horizontal shear is similar. However, it is not hard to show that $\U_H\in\mbox{span}\left(\U_A\right)$ for all possible $ v,h\in\R$. Indeed it can be shown that $\mbox{span}\Big\{\cos^2\theta \hat s, \sin^2\theta \hat s, \sin(2\theta)\hat s\Big \} = \mbox{span}\Big\{\cos^2\theta \hat s, \sin^2\theta \hat s, \{(v^2\sin^2\theta+v\sin2\theta)\hat{s}\}_{v\in\R}, \{(h^2\cos^2\theta+h\sin2\theta)\hat{s}\}_{h\in \R}\Big\} = \mbox{span}\Big\{\cos^2\theta \hat s, \sin^2\theta \hat s, (v_0^2\sin^2\theta+v_0\sin2\theta)\hat{s}, (h_0^2\cos^2\theta+h_0\sin2\theta)\hat{s}\Big\}$ for some fixed $v_0,h_0$. Therefore, in our numerical implementation choosing small fixed $v_0,h_0$ in for $\U_H$ (and hence for $\U_A$) will not change the subspace $\Vk_A$.}.
Next, we orthogonalize $\left\{ \hat{s}_1^{(k)}, \hat{s}_2^{(k)}, \cdots  \right \}\cup\U_A$ to obtain the set of basis vectors $\left\{b_1^{(k)},b_2^{(k)},\cdots\right\}$, which spans the space $\Vk_A$. We then use the basis vectors to define the following matrix:
\begin{align}
    B^{(k)} = \begin{bmatrix}b^{(k)}_1, b^{(k)}_2, \cdots \end{bmatrix}
\end{align}
The proposed training algorithm is outlined as in table~\ref{table:training_algo}.


\begin{table}
\centering
\normalsize
\caption{Training algorithm}
\label{table:training_algo}
\begin{tabular}{ll}
\hline
{\bf{Algorithm:}} Training procedure of the proposed method\\ \hline
{\bf{Input:}} Training images $\{s_1^{(k)},$ $s_2^{(k)},$ $\cdots\}$.\\
{\bf{Output:}} The matrix of basis vectors $B^{(k)}$.\\
{\bf{for}} each class $k$:\\
\hspace{3mm} -- Transform training data to obtain $\left\{ \hat{s}_1^{(k)}, \hat{s}_2^{(k)}, \cdots  \right \}$.\\
\hspace{3mm} -- Orthogonalize $\left\{ \hat{s}_1^{(k)}, \hat{s}_2^{(k)}, \cdots  \right \}\cup\U_A$ to obtain the 
\\\hspace{3mm}  set of basis vectors $\left\{b_1^{(k)},b_2^{(k)},\cdots\right\}$, which spans the 
\\\hspace{3mm}  space $\Vk_A$ (see equation~\eqref{eq:transscal}).\\
\hspace{3mm} -- Use the output of orthogonalization procedure to
\\\hspace{3mm}  define the matrix $B^{(k)}$ containing the basis vectors in 
\\\hspace{3mm}  its columns as follows: $B^{(k)} = \begin{bmatrix}b^{(k)}_1, b^{(k)}_2, \cdots \end{bmatrix}$.\\
\hline
\end{tabular}
\vspace{-1em}
\end{table}

\subsection{Testing algorithm}
\label{section_test}
The testing procedure consists of applying the R-CDT transform followed by a nearest subspace search in the R-CDT space. Let us consider a test image $s$ whose class is to be predicted by the classification method described above. As a first step, we apply the R-CDT on $s$ to obtain the transform space representation $\hat{s}$. We then estimate the distance $d^{2}(\mathbb{P}_\alpha\hat{s}, \Vk_A)\sim \| \mathbb{P}_\alpha\hat{s} - B^{(k)}{B^{(k)}}^T\mathbb{P}_\alpha\hat{s} \|^2$. Here, $\mathbb{P}_\alpha$ is a fixed permutation matrix that causes circular translation in $\theta$ by $\alpha\in[0,\pi]$. Note that $B^{(k)}{B^{(k)}}^T$ is an orthogonal projection matrix onto the space generated by the span of the columns of $B^{(k)}$ (which form an orthogonal basis). The class of $\hat{s}$ is then estimated to be
\begin{equation}
    \arg \min_k\min_\alpha \| \mathbb{P}_\alpha\hat{s} - A^{(k)}\mathbb{P}_\alpha \hat{s} \|^2;~~~~\mbox{where, } A^{(k)}=B^{(k)}{B^{(k)}}^T.\nonumber
\end{equation}
\section{Results}

\begin{figure}[t]
\begin{center}
  \includegraphics[width=1.0\linewidth]{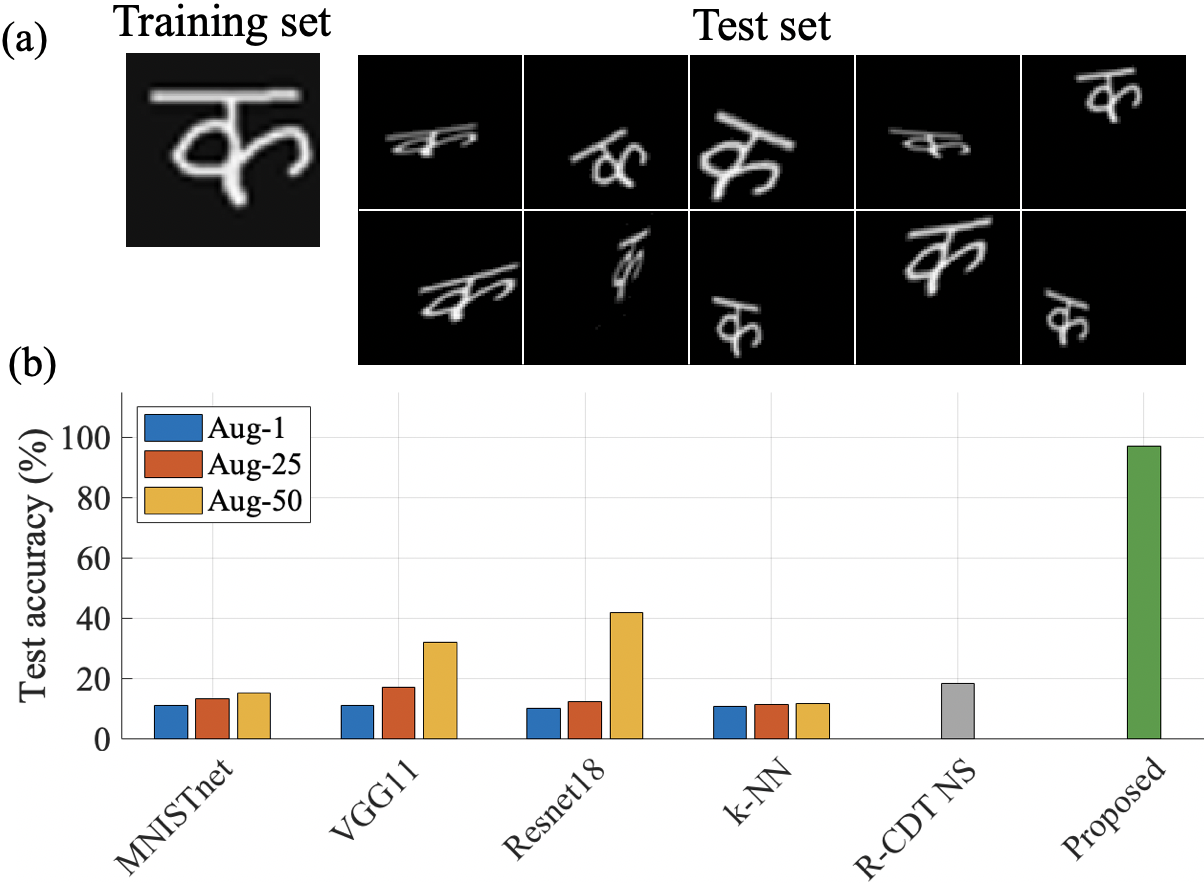}
\end{center}
  \caption{The accuracy of the methods on the synthetic dataset. (a)~Training and test sets of a random class of the synthetic dataset. (b)~The percentage test accuracy of methods. Aug-$1$, Aug-$25$, and Aug-$50$ indicate that the corresponding methods were trained using both original and augmented set where the sizes of the augmented set were $1$, $25$, and $50$ times the size of the original training set, respectively. The R-CDT-NS and the proposed method did not use any augmented images.}
\label{fig:rf01}
\end{figure}

\subsection{Simulated experiment}
We compared the proposed method with conventional classification methods with respect to classification accuracy, data efficiency, computational efficiency, and out-of-distribution robustness. In this respect, we have identified four state-of-the-art methods with data augmentation: MNISTnet \cite{paszke2019pytorch} (a shallow CNN model based on PyTorch's official example), the standard VGG11 model \cite{simonyan2014very}, the standard Resnet18 model \cite{he2016deep}, and the standard k-nearest neighbors (kNN) classifier model \cite{kramer2013k}. The CNN-based methods were implemented using Adam optimizer \cite{kingma2014adam} with $50$ epochs and a learning rate of $0.0005$. Singular value decomposition was used to obtain the basis vectors of the proposed method. The least possible number of basis vectors was chosen so that the sum of variances explained by the selected basis vectors in any class is at least $99\%$ of the total variance in that class, and subspaces corresponding to all classes have the same dimensionality. All methods saw the same set of training and test images.

\begin{figure*}[!hbt]
\begin{center}
\includegraphics[width=1.0\linewidth]{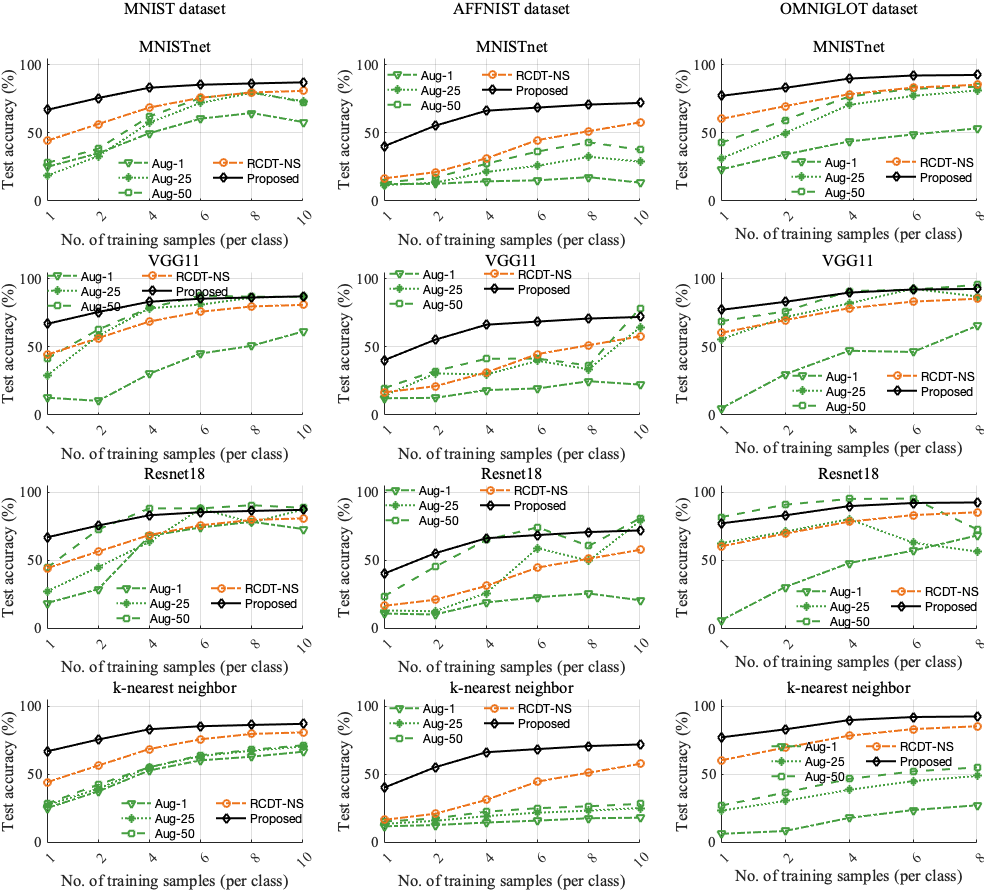}
  \caption{The accuracy of the methods as a function of the number of training samples on the MNIST, AFFNIST, and OMNIGLOT datasets. Aug-$1$, Aug-$25$, and Aug-$50$ indicate that the corresponding methods were trained using both original and augmented set where the sizes of the augmented set were $1$, $25$, and $50$ times the size of the original training set, respectively. The R-CDT-NS and the proposed method did not use any augmented images.}
  \label{fig:rf02}
  \end{center}
\end{figure*}
\begin{figure}[!hbt]
\begin{center}
\includegraphics[width=1.0\linewidth]{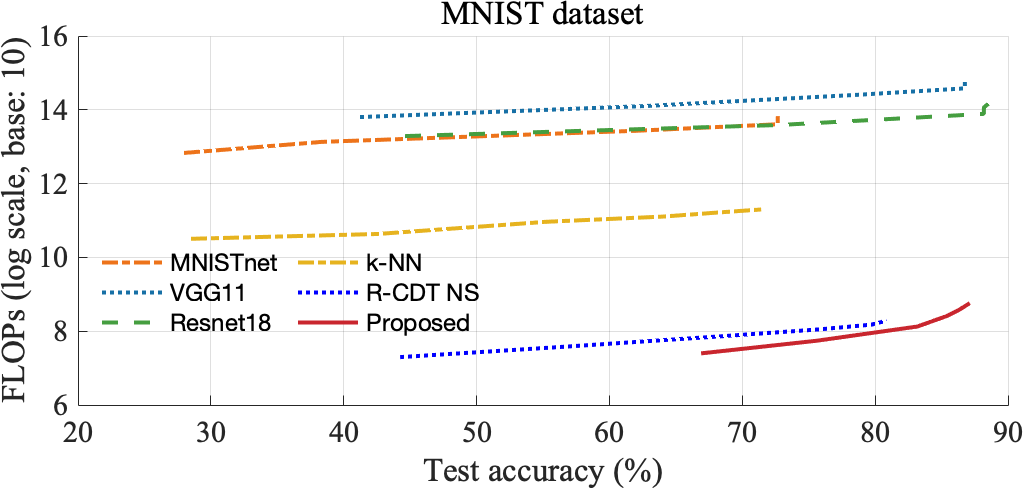}
\end{center}
  \caption{The computational complexity of the methods as measured by the total number of floating-point operations (FLOPs) to attain a particular test accuracy in the MNIST dataset.}
\label{fig:rf03}
\end{figure}

\begin{table*}[]
\centering
\caption{Accuracy of the methods (\%) on images with complex foregrounds}
\label{table:complex_object}
\begin{adjustbox}{width=14cm,center}
\begin{tabular}{lcccccc}
\hline
               & \multicolumn{6}{c}{Training set size (per class) = 1}                               \\
               & MNISTnet      & VGG11         & Resnet18      & k-NN          & R-CDT NS & Proposed \\ 
               & (Aug-1 / Aug-50)  & (Aug-1 / Aug-50)   & (Aug-1 / Aug-50)   & (Aug-1 / Aug-50)  & (Aug-0) & (Aug-0) \\ \hline
Brain MRI & $48.70$ / $51.30$ & $49.60$ / $56.10$ & $49.10$ / $54.00$ & $50.50$ / $50.70$ & $48.70$      & $\bf{57.20}$ \\
Sign Lang & $68.97$ / $74.29$ & $42.53$ / $46.58$ & $40.78$ / $57.78$ & $30.85$ / $32.65$ & $83.00$      & $\bf{87.35}$ \\
OAM (reg) & $5.25$ / $27.18$  & $3.02$ / $16.56$  & $5.77$ / $45.98$  & $17.31$ / $19.40$ & $80.58$      & $\bf{81.51}$ \\
OAM (out) & $4.86$ / $28.21$  & $3.87$ / $15.73$  & $5.28$ / $45.85$  & $20.70$ / $21.88$ & $\bf{85.55}$ & $85.20$      \\
FMNIST    & $34.02$ / $32.68$ & $33.14$ / $33.69$ & $26.45$ / $34.26$ & $25.90$ / $35.52$ & $34.78$      & $\bf{57.90}$ \\ \hline
               & \multicolumn{6}{c}{Training set size (per class) = 5}                               \\
               & MNISTnet      & VGG11         & Resnet18      & k-NN          & R-CDT NS & Proposed \\ 
               & (Aug-1 / Aug-50)  & (Aug-1 / Aug-50)   & (Aug-1 / Aug-50)   & (Aug-1 / Aug-50)  & (Aug-0) & (Aug-0) \\ \hline
Brain MRI & $54.80$ / $55.40$ & $52.80$ / $52.00$ & $50.60$ / $56.30$ & $47.20$ / $49.40$ & $47.80$ & $\bf{62.20}$ \\
Sign Lang & $91.05$ / $91.59$ & $44.66$ / $79.93$ & $47.49$ / $91.46$ & $92.01$ / $92.40$ & $93.21$ & $\bf{96.10}$ \\
OAM (reg) & $43.61$ / $68.48$ & $29.37$ / $26.16$ & $72.62$ / $81.84$ & $34.03$ / $38.92$ & $92.69$ & $\bf{93.71}$ \\
OAM (out) & $38.78$ / $56.24$ & $25.41$ / $41.32$ & $67.70$ / $58.67$ & $37.69$ / $41.83$ & $91.20$ & $\bf{91.44}$ \\
FMNIST    & $43.97$ / $50.92$ & $31.87$ / $65.46$ & $31.67$ / $65.22$ & $38.69$ / $42.95$ & $54.31$ & $\bf{82.64}$  \\ \hline
               & \multicolumn{6}{c}{Training set size (per class) = 10}                               \\
               & MNISTnet      & VGG11         & Resnet18      & k-NN          & R-CDT NS & Proposed \\ 
               & (Aug-1 / Aug-50)  & (Aug-1 / Aug-50)   & (Aug-1 / Aug-50)   & (Aug-1 / Aug-50)  & (Aug-0) & (Aug-0) \\ \hline
               & MNISTnet          & VGG11             & Resnet18                    & k-NN              & R-CDT NS     & Proposed     \\
Brain MRI & $50.70$ / $54.10$ & $50.40$ / $57.00$ & $50.60$ / $60.00$ & $49.00$ / $50.30$ & $52.10$ & $\bf{62.30}$ \\
Sign Lang & $83.82$ / $88.52$ & $43.30$ / $75.52$ & $39.61$ / $87.87$ & $95.77$ / $95.74$ & $97.47$ & $\bf{98.26}$ \\
OAM (reg) & $54.41$ / $79.40$ & $69.38$ / $87.85$ & $88.50$ / $95.43$ & $45.46$ / $51.23$ & $95.82$ & $\bf{97.34}$ \\
OAM (out) & $44.27$ / $60.72$ & $57.28$ / $75.06$ & $72.38$ / $79.27$ & $46.83$ / $51.28$ & $92.76$ & $\bf{94.16}$ \\
FMNIST    & $37.95$ / $58.62$ & $42.42$ / $74.49$ & $37.45$ / $78.38$ & $46.55$ / $49.66$ & $75.68$ & $\bf{86.14}$  \\ \hline
\end{tabular}
\end{adjustbox}
\end{table*}

To evaluate the comparative performance of the methods, we selected the following datasets: MNIST, Affine-MNIST, OMNIGLOT, Brain MRI, Sign language, OAM (under the regular and the out-of-distribution setup), and FMNIST datasets. The handwritten character images of the MNIST and OMNIGLOT datasets were selected from \cite{lecun1998gradient} and \cite{lake2015human}, respectively. The 2D images from the middle slices of the 3D MRI data of the Brain MRI dataset and normalized HOGgles images of hand gestures of the Sign language dataset were collected from \cite{shifat2021radon}. The fashion object images (trouser, pullover, bag, ankle boot) of the FMNIST dataset were collected from \cite{xiao2017fashion}. The optical communication images (orbital angular momentum beam patterns) of the OAM dataset under the influence of various atmospheric turbulence levels were collected from \cite{nichols2018transport}. We tested the methods on the OAM dataset under two experimental setups: the regular setup, where the training and test sets contain images at the same turbulence level, and the out-of-distribution setup, where the training and test sets contain images at different turbulence levels. We randomly selected $8$ images from the OMNIGLOT dataset for training, and the rest were used for testing. The Affine-MNIST dataset was created using random Affine transformations (translation, isotropic/anisotropic scaling, shear, rotation) to both training and test sets of the MNIST dataset. To increase the classification complexity, random affine transformations were also added with the Brain MRI and FMNIST datasets. We also created a synthetic dataset using randomly selected ten classes of the OMNIGLOT dataset \cite{lake2015human}. The training set of the synthetic dataset contains a randomly selected $1$ image per class, and the test set contains $200$ instances of the same image but observed under random affine transformations (translation, isotropic/anisotropic scaling, shear, and rotation). Fig.~\ref{fig:rf01}(a) illustrates the single image of the training set and a few sample images of the test set of a random class of the synthetic dataset.

\subsection{Effectiveness and data-efficiency}
The proposed method provides an effective and data-efficient solution in the synthetic dataset (see Fig.~\ref{fig:rf01}(b)). Note that the MNISTnet, VGG11, Resnet18, and kNN models were trained using the original training set in addition to the augmented training set generated from the original training set. The sizes of the augmented training set used were $1$, $25$, and $50$ times the size of the original training set. We also emphasize that the R-CDT-NS method and the proposed method did not use augmented images; they were trained only using the original training set. The proposed method provides test accuracy close to $100\%$ using no augmented data, whereas the other methods reach accuracy up to $40\%$ using $50$ times more augmented images (see Fig.~\ref{fig:rf01}(b)).

The classification accuracy, for different training set sizes, for the real datasets are shown in Fig.~\ref{fig:rf02} and Table~\ref{table:complex_object}. Note that, in the real datasets also, the MNISTnet, VGG11, Resnet18, and kNN models were trained using the original training set in addition to the augmented training set where the sizes of the augmented training set used were $1$, $25$, and $50$ times the size of the original training set. Also, note that the R-CDT-NS method and the proposed method were implemented without using any augmented images. Results in the real datasets show that the proposed method provides accuracy better than or equivalent to the other methods without using any augmented images, where the other methods used up to $50$ times more augmented images (see Fig.~\ref{fig:rf02}). The proposed method outperforms the other methods by a more significant margin at the low-training sample end. Increasing the augmentation size improves the performance of the other methods, but it significantly increases the other methods' computational burden, as will be clarified in the next section.

\begin{figure*}[!hbt]
\begin{center}
\includegraphics[width=1.0\linewidth]{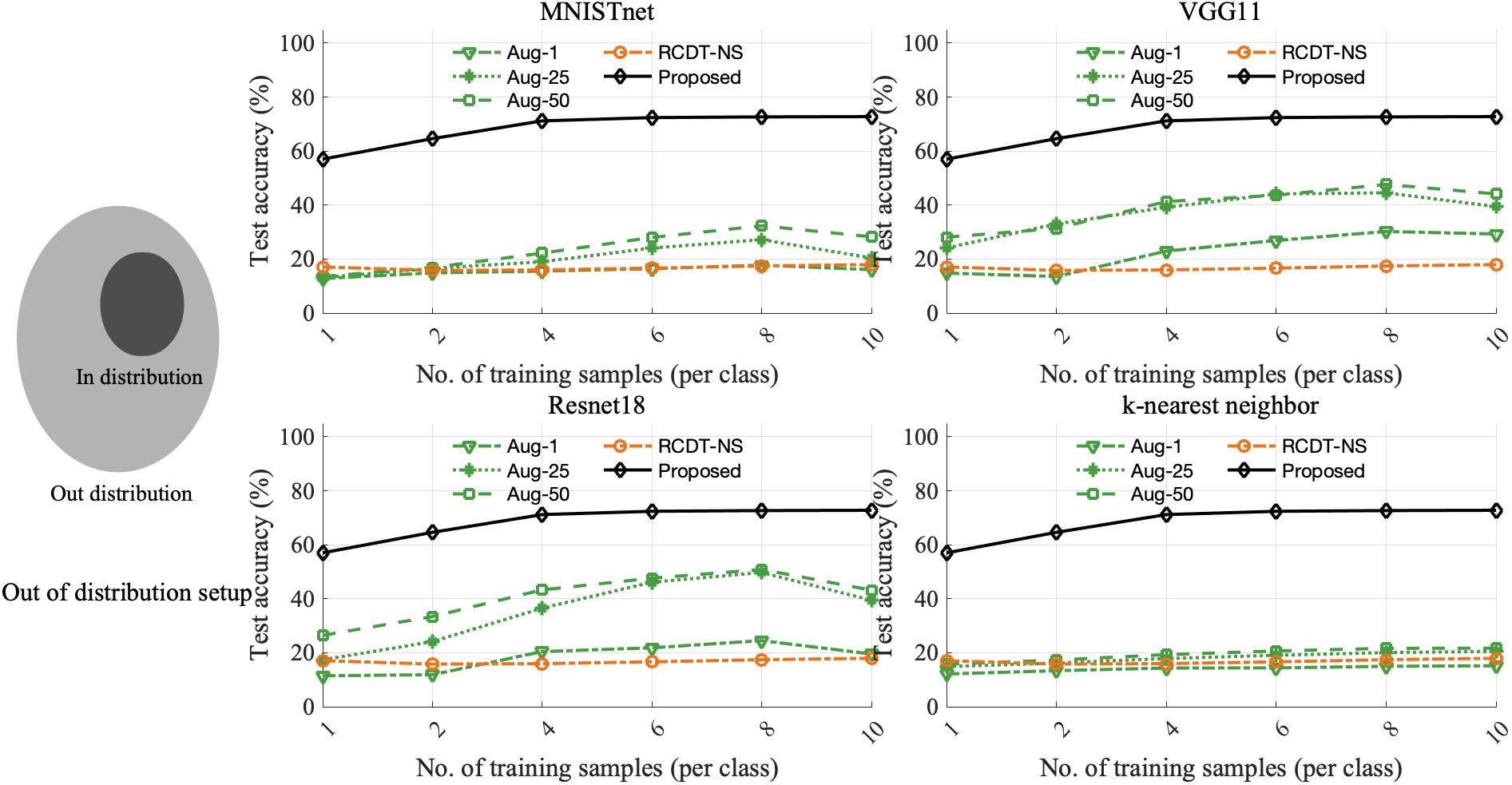}
\end{center}
  \caption{Experimental results under the out-of-distribution setup, which is characterized by the disjoint training (`in distribution') and test (`out distribution') sets containing different sets of magnitudes of the spatial transformations (see the left panel). The percentage test accuracy values of different methods are measured as a function of the number of training images per class.}
\label{fig:rf04}
\end{figure*}

\subsection{Out-of-distribution robustness}
To compare the effectiveness of the proposed method under the out-of-distribution setting, we generated a gap between the training and test sets with respect to the magnitudes of the deformations. Formally, if $\G_{in}\subset\G$ denotes the deformation set of the `in-distribution', then $\G_{out}=\G \backslash \G_{in}$ was used as the deformation set for the `out-distribution' (see Fig.~\ref{fig:rf04}). Then we trained the models using the `in-distribution' data and tested using the `out-distribution' data. We performed two out-of-distribution experiments. In one experiment, we used the MNIST dataset as the `in-distribution' training set and the Affine-MNIST dataset as the `out-distribution' test set (see Fig.~\ref{fig:rf04}), and in the other out-of-distribution experiment, we used the OAM images at low turbulence levels as the `in-distribution' training set and those at medium and high turbulence levels as the `out-distribution' test set as in \cite{shifat2021radon} (see Table~\ref{table:complex_object}). The range of deformation magnitudes used in the augmentation set was also chosen to be different from the Affine-MNIST dataset in this out-of-distribution experiment. The results show that the proposed method outperforms the other methods by an even more significant margin under the challenging out-of-distribution setup (see Fig.~\ref{fig:rf04} and Table~\ref{table:complex_object}). Under this setup, the proposed method maintains similar accuracy figures in the Affine-MNIST and OAM test data compared with the standard experimental setup (i.e., Affine-MNIST in Fig.~\ref{fig:rf02} and OAM (regular) in Table~\ref{table:complex_object}). On the other hand, the other methods decline in accuracy significantly under the out-of-distribution setup compared with the standard experimental setup (see Affine-MNIST and OAM results in Figs.~\ref{fig:rf02},~\ref{fig:rf04} and Table~\ref{table:complex_object}).

\subsection{Computational efficiency}
To compare the computational efficiency of the methods, we computed the number of the floating-point operations (FLOPs) \cite{shifat2021radon} in the training phase of the methods (see the FLOPs vs. percentage accuracy results for the MNIST dataset in Fig.~\ref{fig:rf03}). The results show that the proposed method requires up to $6$ orders of magnitude ($1,000,000$ times) less computational cost than the other methods to achieve the same test accuracy. As in the previous experiments, we augmented the training set for the MNISTnet, VGG11, Resnet18, and kNN methods and did not augment the training set for the R-CDT-NS and the proposed method. The size of the augmentation set used in this experiment was $50$ times more than the original training set. The computational complexity of other methods would potentially reduce if the augmentation set size were reduced, but that would also aggravate their classification accuracy (see Fig.~\ref{fig:rf02}).


\section{Discussion}
The results above show that our method's mathematically prescribed invariance encoding technique can reasonably model the specific deformation set under consideration, i.e., the affine set. The proposed method offers high classification accuracy using significantly less training data and computation without explicitly using any augmented images. The method is robust under challenging experimental setups such as out-of-distribution testing cases.

\subsection*{Test accuracy and data efficiency}
Results in synthetic and real data show that, so long as the data at and conform to the generative model stated in equation~\eqref{eq:2dgenerative_model} (or equation~\eqref{eq:2dgenerative_model_im}; also depicted in Fig.~\ref{fig:f00}), the proposed method can classify images with high accuracy without explicitly using any augmented images. While the proposed and RCDT-NS methods were implemented using no augmented images, the other methods we compared to were implemented using low to high numbers of augmented images. The other methods significantly underperform while using a low number of augmented images (see Figs.~\ref{fig:rf01}(b), \ref{fig:rf02}, and \ref{fig:rf04}). The other methods, in some cases, offer equivalent classification accuracy to the proposed method while using a significantly high number of augmented images (see Fig.~\ref{fig:rf02}). However, this approach significantly increases the computational burden of the other methods (see Fig.~\ref{fig:rf03}). Here we emphasize that, though the proposed method was implemented using any data augmentation here, data augmentation could also be used in the proposed method as in the other methods. In that case, the accuracy of the proposed method could potentially improve even further. 

\subsection*{Computational efficiency}
The proposed method provides up to 6 orders of magnitude (1,000,000 times) savings in computational cost (see Fig.~\ref{fig:rf03}) as measured by the number of floating-point operations (FLOPs) over data augmented deep learning alternatives. Such an improvement in computational efficiency could be achieved due to the simple and non-iterative nature of the proposed solution. The MNISTnet, VGG11, Resnet18, and kNN methods require a high number of augmented images to achieve reasonable accuracy figures and require iterative optimization procedures to reach a solution, contributing to high computational costs. Computational costs of these methods could be reduced by reducing the number of augmented images used, but that would also adversely affect their classification accuracy (see Fig.~\ref{fig:rf02}).

\subsection*{Out-of-distribution robustness}
The proposed method maintains high classification accuracy while the accuracy figures of the other methods fall drastically under the challenging out-of-distribution experimental setup (see Fig.~\ref{fig:rf04}). These results suggest that the proposed method provides a better generalization of the underlying data distribution resulting in robust classification performance. The reason for better accuracy under the out-of-distribution setup is that the proposed method does not only learn the deformations present in the given data; it actually learns the underlying data model. More specifically, it learns the type of deformation (such as translation, scaling, shear, and others) present in the data. It thereby can detect the presence of different magnitudes of these deformation types. The type of the deformation can be learned from a very few training samples containing those deformations as well as from the mathematically prescribed invariances proposed in this paper. 

\section{Conclusion}
This paper proposes an enhanced end-to-end classification system with a mathematical framework to attain invariances to a set of given image transformations. The proposed method is pertinent to a specific category of image classification problems where image classes can be thought of being an instance of a template observed under a set of spatial deformations. If these deformations are appropriately modeled as a collection of smooth, one-to-one, and nonlinear transformations (see equations \eqref{eq:2dgenerative_model_im} and \eqref{eq:2dgenerative_model} of the paper), then the image classes become easily separable in the transform space (i.e., the R-CDT space) via the properties mentioned in \cite{shifat2021radon}. These properties also allow for the approximation of image classes as convex subspaces in the R-CDT space, providing a more suitable data model for the nearest subspace method. The resulting classifier can then be expected to provide high accuracy, computational efficiency, and out-of-distribution robustness, as we found in the experiments. A large number of image classification problems can be formulated this way and thus can benefit from our proposed solution.  Heuristically, any classification problem for which one image in a class can be constructed from another by a smooth rearrangement of pixel intensities is an appropriate fit for the generative model. Obvious examples are affine transformations (translation, scaling, shear, etc.). Less obvious examples are distortions to images in an optical communication channel resulting from the influence of a transparent medium (e.g., turbulence, see \cite{nichols2018transport}) or morphological changes in the gray matter of MRI images under the influence of a disease (see \cite{kundu2018discovery}).

The proposed mathematical solution attains high classification accuracy (compared with state-of-the-art end-to-end systems), especially at the low data regime. The method was also demonstrated to significantly improve the computational cost of classification (up to 1,000,000 times reduction in the computational cost can be attained). The method is mathematically coherent, understandable, non-iterative, requires no hyper-parameters to tune, and is simple enough to be implemented without GPU support. However, the proposed method can also be implemented in parallel using a GPU, which should further enhance the method’s efficiency. The method also demonstrated robustness in challenging experimental scenarios, e.g., the out-of-distribution setup. The method performs well under the out-of-distribution setup because it learns the underlying generative model of the image classes. More specifically, the method learns the type of deformations that might have generated the dataset by using very few training examples.

We obtain superior performance by expanding upon the recently published R-CDT-NS classification method \cite{shifat2021radon}, which can also be interpreted as the `nearest' sliced-Wasserstein distance method. The R-CDT-NS method \cite{shifat2021radon} was demonstrated to show equivalent or better classification accuracy at both low and high data regimes. In this paper, we improved upon the performance of the R-CDT-NS method at the low data regime without altering its previously superior performance at the high data regime. The performance improvement at the low data regime was achieved by improving the invariance prescribing framework of the R-CDT-NS method. In the proposed method, we encode invariances with respect to a more complicated deformation set than the previous paper \cite{shifat2021radon}: the affine deformations, i.e., translation, isotropic/anisotropic scaling, horizontal/vertical shear, and rotation in the sliced-Wasserstein space. Though images under the effect of these deformations are challenging to classify in native image space, the R-CDT subspace can capture these variations and thus simplify the associated classification problems. We mathematically derive approximate basis vectors corresponding to these deformations and use them to enhance the R-CDT subspace to encode invariances instead of augmenting the individual training images. As a result, the method can learn a specific deformation type using a few basis vectors without requiring to use thousands of augmented images representing that deformation.

Finally, we note that the method is well-suited for the problems where the data at hand conform to the generative model stated in equation~\eqref{eq:2dgenerative_model}. One example where the data do not follow the generative model is classification problems involving natural unsegmented images (e.g., CIFAR10, imagenet datasets). However, some datasets (such as the OAM and Sign language datasets) contain unsegmented images, and the proposed method still outperforms the other methods in these datasets. In addition, our method can potentially be extended to be suitably applied to natural unsegmented images with more complex backgrounds. However, it would require redefining the problem statement and the generative model. One step forward in this direction is a few recent papers \cite{zhuang2022local,zhang2020deepemd} that consider images as a collection of patches. When an image is considered as a collection of patches, it might be possible to adaptively assign lower weights to the background and discard them automatically. However, these analyses require reformulation of the problem, which we leave to future work. Another potential solution is to use an object detection and segmentation method along with our proposed classification method. We also note that we have used some approximations and assumptions in the derivation for the spanning sets of shear and anisotropic scaling. However, we have shown that for practical purposes, these approximations work reasonably as we have seen in the results provided above.

\section*{Acknowledgments}
This work was supported in part by NIH grant GM130825, and NSF grant 1759802. Authors acknowledge Drs. Watnik and Nichols for the permission to use the OAM dataset shown in Table~\ref{table:complex_object}.

{\small
\bibliographystyle{elsarticle-num}
\bibliography{egbib}
}

\clearpage
\newpage

\section*{Appendix A}

\subsection*{Anisotropic scaling}
Consider two functions $s_g(x,y)$ and $s(x,y)$ such that
\begin{align}
\label{eqn:eq1}
    s_g(x,y)=\frac{1}{ab}s(\frac{x}{a},\frac{y}{b}),
\end{align}
for some $a,b>0$.
By definition of the Radon transform, we have
\begin{align}
    \tilde{s}_g(t,\theta)=\frac{1}{ab}\int_{-\infty}^{\infty}\int_{-\infty}^{\infty}s(\frac{x}{a},\frac{y}{b})\delta(t-x\cos\theta-y\sin\theta)dxdy.
\end{align}
Applying the change of variables formula with $x^{\prime} = \frac{x}{a}$ and $y^{\prime} = \frac{y}{b}$, we have that 
\begin{align}
    \tilde{s}_g(t,\theta)=\int_{-\infty}^{\infty}\int_{-\infty}^{\infty}s(x^{\prime},y^{\prime})\delta(t-ax^{\prime}\cos\theta-by^{\prime}\sin\theta)dx^{\prime}dy^{\prime}.
\end{align}
Using the co-area formula and letting $\gamma=\sqrt{a^2\cos^2\theta+b^2\sin^2\theta}$, we have that

\begin{align}
    \tilde{s}_g(t,\theta)&=\frac{1}{\gamma}\int_{-\infty}^{\infty}\int_{-\infty}^{\infty}s(x^{\prime},y^{\prime})\delta(\frac{t}{\gamma}-x^{\prime}\frac{a\cos\theta}{\gamma}-y^{\prime}\frac{b\sin\theta}{\gamma})dx^{\prime}dy^{\prime}\\
    &= \frac{1}{\gamma}\int_{-\infty}^{\infty}\int_{-\infty}^{\infty}s(x^{\prime},y^{\prime})\delta(\frac{t}{\gamma}-x^{\prime}\cos\theta^{\prime}-y^{\prime}\sin\theta^{\prime}) dx^{\prime}dy^{\prime},
\end{align}
where $\theta^\prime=\tan^{-1}\left(\frac{b}{a}\tan\theta\right)$.
Hence
\begin{equation}
    \tilde{s}_g(t,\theta)=\frac{1}{\gamma}\tilde{s}\left(\frac{t}{\gamma},\theta^\prime\right)\end{equation}
Applying the scaling property of R-CDT
\begin{align}
\label{aniso_gam}
    \hat{s}_g(t,\theta)=\gamma\hat{s}(t,\theta^\prime)
\end{align}

We are interested in the scenario where $a\approx b$. Let $\frac{b}{a}= 1+ \alpha$ with some $\alpha\approx 0$. For illustration purposes, we assume without loss of generality  $\theta \in [0,\frac{\pi}{2})$ and $a\leq b$ (i.e., $\alpha\geq 0$) in the following derivations. Other cases are similar.  Using Taylor's formula for $\tan^{-1}(x)$ around $x= \tan \theta$  , we have that
\begin{align}
	\theta^{\prime} &=\tan^{-1}\left(\tan\theta+\alpha \tan \theta \right)\\
	&= \theta + \frac{\alpha\tan \theta}{1+\tan^2\theta} - \frac{\xi}{(1+\xi^2)^2}(\alpha \tan\theta)^2,
\end{align}
where $\xi \in [\tan \theta, (1+\alpha)\tan\theta]$.
Since $0\leq\frac{|\xi|}{1+{\xi}^2}\leq \frac{1}{2}$ and $\frac{1}{1+\xi^2}\leq \frac{1}{1+\tan^2\theta}$ for $\xi \geq \tan\theta\geq 0$, we have that 
\begin{equation}
	|\theta^{\prime}-\theta|\leq (\alpha\sin\theta\cos\theta +\frac{1}{2}\alpha^2\sin^2\theta).
\end{equation}
With the observation that $|\sin\theta\cos\theta| \leq \frac{1}{2}$, it is also easy to derive from above a bound of $|\theta^{\prime}-\theta|$ independent of $\theta$:
\begin{equation}
	|\theta^{\prime}-\theta|\leq \frac{1}{2}(\alpha +\alpha^2).
\end{equation}

In practice, we choose $45$ uniform angles between $0$ and $\pi$, the difference of consecutive angles are $\frac{\pi}{45}\approx 0.07$. Choosing $
\alpha$ between $0$ and $.12$ guarantee that $|\theta^{\prime}-\theta|\leq .068$. Similarly one can show that for $-1<\alpha<0$, 
\begin{equation}
	|\theta^{\prime}-\theta|\leq |\alpha\sin\theta\cos\theta|+\frac{1}{2}\frac{\alpha^2\tan^2\theta}{1+(1+\alpha)^2\tan^2\theta}) \leq \frac{1}{2}(|\alpha| + \frac{\alpha^2}{(1+\alpha)^2}).
\end{equation}
Choosing $\alpha\in [-.12,0)$, we have that $|\theta^{\prime}-\theta|\leq .07$. 

In summary, with $.88\leq\frac{b}{a}\leq 1.12$, the difference between $\theta^{\prime}$ and $\theta$ is smaller than the numerical difference of consecutive angles used in Radon transform computation in our algorithm. Therefore, using equation~\eqref{aniso_gam}, we have that
\begin{align}
\label{aniso_gam2}
    \hat{s}_g(t,\theta)=\gamma\hat{s}(t,\theta^\prime) \approx \gamma\hat{s}(t,\theta)
\end{align}

Next we show an approximation for $\gamma$ for $a\leq b$. Observing that $\gamma = a\cos\theta \sqrt{1+(1+\alpha)^2\tan^2\theta} =  a\cos\theta\\ \sqrt{1+\tan^2\theta +(2\alpha+\alpha^2)\tan^2\theta}$ and using Taylor's formula for $\sqrt{x}$ around $x = 1+\tan^2\theta$, we have that
\begin{equation}\label{tayloralpha}
	\gamma = a\cos\theta \big(\sqrt{1+\tan^2\theta}+ \frac{(2\alpha+\alpha^2)\tan^2\theta}{2\sqrt{1+\tan^2\theta}}- \frac{(2\alpha+\alpha^2)^2\tan^4\theta}{8(\xi)^{3/2}}\big),
\end{equation} 
where $\xi \in [1+\tan^2\theta, 1+(1+\alpha)^2\tan^2\theta]$. Observing that ${\sqrt{1+\tan^2\theta}}=\frac{1}{\cos \theta}$ and $\frac{1}{\xi}\leq \frac{1}{1+\tan^2\theta}$ for $\xi \in [1+\tan^2\theta, (1+\alpha)^2\tan^2\theta]$, we obtain that
\begin{align}
	|\gamma - a| &\leq a\cos\theta \Big((\alpha +\frac{\alpha^2}{2})\cos \theta\tan^2\theta +\frac{(2\alpha+\alpha^2)^2}{8}\cos^3\theta\tan^4\theta\Big)\\
	& = a(\alpha +\frac{\alpha^2}{2})\sin^2\theta + a\frac{(2\alpha+\alpha^2)^2}{8}\sin^4\theta.
\end{align}
Ignoring higher order terms of $\alpha$ in \eqref{tayloralpha}, we have the following approximation
\begin{equation}
	\gamma = a+\alpha a\sin^2\theta+\bigo(\alpha^2) = (1+\alpha\sin^2\theta)a + \bigo(\alpha^2).
\end{equation}
Analogously, if $a>b$, we let $\frac{a}{b}= 1+\beta$ for some $\beta>0$ and by similar arguments we have that
\begin{equation}\label{taylorbeta}
	|\gamma - b|\leq b(\beta +\frac{\beta^2}{2})\cos^2\theta + b\frac{(2\beta+\beta^2)^2}{8}\cos^4\theta.
\end{equation}
Ignoring higher order terms of $\beta$ in \eqref{taylorbeta}, we have the following approximation
\begin{equation}
	\gamma = b(1+\beta\cos^2\theta) +  \bigo(\beta^2) .
\end{equation}
Hence $\hat{s}_g(t,\theta)\approx \gamma \hat s(t,\theta) = a\hat s(t,\theta)+ \alpha a\sin^2\theta\hat s(t,\theta) +\bigo(\alpha^2)$ for $a<b$ and $\hat{s}_g(t,\theta)\approx \gamma \hat s(t,\theta) = b\hat s(t,\theta)+ \beta b\cos^2\theta\hat s(t,\theta)+\bigo(\beta^2)$ for $a>b$.
In summary, to model anisotropic scalings of $s$ as in equation~\eqref{eqn:eq1} with $a\approx b$, we add the following additional spanning set $\hat E = \{\cos^2\theta \hat s, \sin^2\theta \hat s\}$ as enrichment to the training subspace in the transform space.

\section*{Appendix B}

\subsection*{Shear-horizontal}
Consider two functions $s_{g_1}(x,y)$ and $s(x,y)$ such that for some $h\in\R$, 
\begin{align}
\label{eqn:eq2}
    s_{g_1}(x,y)=s(x-hy,y).
\end{align}
By definition of the Radon transform, we have
\begin{align}
    \tilde{s}_{g_1}(t,\theta)=\int_{-\infty}^{\infty}\int_{-\infty}^{\infty}s(x-hy,y)\delta(t-x\cos\theta-y\sin\theta)dxdy
\end{align}
Applying the change of variables formula with $x^\prime=x-hy, y^{\prime} = y$, we have that
\begin{align}
    &\tilde{s}_{g_1}(t,\theta)\nonumber\\
    &=\int_{-\infty}^{\infty}\int_{-\infty}^{\infty}s(x^\prime,y^{\prime})\delta(t-x^\prime\cos\theta-hy^{\prime}\cos\theta-y^{\prime}\sin\theta)dx^\prime dy^{\prime}\nonumber\\
    &=\int_{-\infty}^{\infty}\int_{-\infty}^{\infty}s(x^\prime,y^{\prime})\delta(t-x^\prime\cos\theta-y^{\prime}(\sin\theta+h\cos\theta))dx^\prime dy^{\prime}.
\end{align}
Using the co-area formula and the scaling properties of the Dirac delta function and letting $\gamma=\sqrt{1+h^2\cos^2\theta+h\sin(2\theta)}$, we have
\begin{align}
    &\tilde{s}_{g_1}(t,\theta) \nonumber\\ &=\int_{-\infty}^{\infty}\int_{-\infty}^{\infty}\frac{1}{\gamma} s(x^\prime,y^{\prime})\delta(\frac{t}{\gamma}-\frac{x^\prime\cos\theta-y^{\prime}(\sin\theta+h\cos\theta))}{\gamma}dx^\prime dy^{\prime}.
\end{align}

Let $\theta^{\prime} = \tan^{-1}(\frac{\sin\theta+h\cos\theta}{\cos\theta})=  \tan^{-1}(\tan \theta+h)$, it is not hard to check that $ \cos \theta^{\prime} = \frac{\cos \theta}{\gamma}$ and $\sin\theta^{\prime}= \frac{\sin\theta+h\cos\theta}{\gamma}$. Hence
\begin{align}
    \tilde{s}_{g_1}(t,\theta)  &=\int_{-\infty}^{\infty}\int_{-\infty}^{\infty}\frac{1}{\gamma} s(x^\prime,y^{\prime})\delta(\frac{t}{\gamma}-x^\prime\cos\theta^{\prime}-y^{\prime}\sin\theta^{\prime})dx^\prime dy^{\prime}.
\end{align}
By the definition of Radon transform, we see that

\begin{align}
    \tilde{s}_{g_1}(t,\theta)=\frac{1}{\gamma}\tilde{s}\left(\frac{t}{\gamma},\theta^\prime\right);&~\gamma=\sqrt{1+h^2\cos^2\theta+h\sin(2\theta)},\nonumber\\
    &~\theta^\prime=\tan^{-1}\left(\tan\theta+h\right).
\end{align}
Applying the scaling property of R-CDT, we have that
\begin{align}
    \label{hor_shear_gam}
    \hat{s}_{g_1}(t,\theta)=\gamma\hat{s}(t,\theta^\prime).
\end{align}

We are interested in the scenario where $h\approx 0$.  For illustration purposes, we assume without loss of generality  $\theta \in [0,\frac{\pi}{2})$ and $h\geq 0$ in the following derivations. Other cases are similar.  Using Taylor's formula for $\tan^{-1}(x)$ around $x= \tan \theta$  , we have that
$\tan^{-1}(x)$ around $x= \tan \theta$  , we have that
\begin{align}
	\theta^{\prime} &=\tan^{-1}\left(\tan\theta+h\right)\\
	&= \theta + \frac{h}{1+\tan^2\theta} - \frac{\xi h^2}{(1+\xi^2)^2},
\end{align}
where $\xi \in [\tan \theta, \tan\theta+h]$. Since $0\leq\frac{|\xi|}{1+{\xi}^2}\leq \frac{1}{2}$ and $\frac{1}{1+\xi^2}\leq \frac{1}{1+\tan^2\theta}$ for $\xi \geq \tan\theta\geq 0$, we have that 
\begin{equation}
	|\theta^{\prime}-\theta|\leq (h +\frac{1}{2}h^2)\cos^2\theta.
\end{equation}
With the observation that $|\cos\theta| \leq 1$, it is easy to see that \begin{equation}
	|\theta^{\prime}-\theta|\leq  h +\frac{1}{2}h^2.
\end{equation}

In practice, we choose $45$ uniform angles between $0$ and $\pi$, choosing $
h$ between $0$ and $.067$ guarantee that $|\theta^{\prime}-\theta|\leq \frac{\pi}{45}$. Similarly one can show that for $h<0$, 
\begin{equation}
	|\theta^{\prime}-\theta|\leq |h| + h^2.
\end{equation}
It is easy to see that if $-.065\leq h<0$, then $|\theta^{\prime}-\theta|\leq \frac{\pi}{45}$. In summary, with $|h|\leq .065$, the difference between $\theta^{\prime}$ and $\theta$ is smaller than the numerical difference of consecutive angles used in Radon transform computation in our algorithm. Therefore, using equation~\eqref{hor_shear_gam}, we have
\begin{align}
    \label{hor_shear_gam2}
    \hat{s}_{g_1}(t,\theta)=\gamma\hat{s}(t,\theta^\prime)\approx\gamma\hat{s}(t,\theta).
\end{align}

Next we show an approximation for $\gamma$. Apply Taylor's formula for $\sqrt{x}$ around $x = 1$ to $\gamma$, we have that
\begin{align}\label{tayloralpha}
	\gamma &= \sqrt{1+h^2\cos^2\theta+h\sin(2\theta)} \nonumber\\
	&= 1+ \frac{1}{2}\big(h^2\cos^2\theta+h\sin(2\theta)\big)-\frac{\big(h^2\cos^2\theta+h\sin(2\theta)\big)^2}{8\xi^{3/2}},
\end{align} 
where $\xi \in [1, 1+h^2\cos^2\theta+h\sin(2\theta)]$. Ignoring higher order terms we have that for $h\geq 0$
\begin{equation}
	\gamma = 1+ \frac{1}{2}\big(h\sin(2\theta) +h^2\cos^2\theta\big)+ \bigo(h^2),
\end{equation}
with  
\begin{equation}
	|\gamma - 1| \leq\frac{1}{2}(h+h^2)+\frac{1}{8}(h+h^2)^2.
\end{equation}
Hence $\hat{s}_{g_1}(t,\theta)\approx \gamma \hat s(t,\theta) = \hat s(t,\theta)+ \frac{1}{2}(h\sin(2\theta)+h^2\cos^2\theta)\hat s(t,\theta)+\bigo(h^2)$. 
In summary, to model small horizontal shearing of $s$ as in \eqref{eqn:eq2} with $h\approx 0$, we add the following additional spanning set $\hat E =\{ \big(h^2\cos^2\theta+h\sin(2\theta) \big)\hat s\}$ (for small $h$) as enrichment to the training subspace in the transform space.
\subsection*{Shear-vertical}
Consider the functions $s_{g_2}(x,y)$ and $s(x,y)$ such that for some $v\in\R$
\begin{align}
\label{eqn:eq3}
    s_{g_2}(x,y)=s(x,y-vx).
\end{align}

By similar arguments as above, we have that
\begin{align}
    \tilde{s}_{g_2}(t,\theta)=\frac{1}{\gamma}\tilde{s}\left(\frac{t}{\gamma},\theta^\prime\right);~&\gamma=\sqrt{1+v^2\sin^2\theta+v\sin(2\theta)},\nonumber\\
    ~&\theta^\prime=\cot^{-1}\left(\cot\theta+v\right)
\end{align}
Applying the scaling property of R-CDT, we have that
\begin{align}
    \label{ver_shear_gam}
    \hat{s}_{g_2}(t,\theta)=\gamma\hat{s}(t,\theta^\prime).
\end{align}

We are interested in the scenario where $v\approx 0$.  For illustration purposes, we assume without loss of generality  $\theta \in [0,\frac{\pi}{2})$ and $v\geq 0$ in the following derivations. Other cases are similar.  Using Taylor's formula for $\tan^{-1}(x)$ around $x= \tan \theta$  , we have that
$\cot^{-1}(x)$ around $x= \cot \theta$  , we have that
\begin{align}
	\theta^{\prime} &=\cot^{-1}\left(\cot\theta+v\right)\\
	&= \theta - \frac{v}{1+\cot^2\theta} +\frac{\xi v^2}{(1+\xi^2)^2},
\end{align}
where $\xi \in [\cot \theta, \cot\theta+v]$. Since $0\leq\frac{|\xi|}{1+{\xi}^2}\leq \frac{1}{2}$ and $\frac{1}{1+\xi^2}\leq \frac{1}{1+\cot^2\theta}$ for $\xi \geq \cot\theta\geq 0$, we have that 
\begin{equation}
	|\theta^{\prime}-\theta|\leq (v +\frac{1}{2}v^2)\sin^2\theta.
\end{equation}
With the observation that $|\cos\theta| \leq 1$, it is easy to see that \begin{equation}
	|\theta^{\prime}-\theta|\leq  v +\frac{1}{2}v^2.
\end{equation}

In practice, we choose $45$ uniform angles between $0$ and $\pi$, choosing $
h$ between $0$ and $.067$ guarantee that $|\theta^{\prime}-\theta|\leq \frac{\pi}{45}$. Similarly one can show that for $h<0$, 
\begin{equation}
	|\theta^{\prime}-\theta|\leq |v| + v^2.
\end{equation}
It is easy to see that if $-.065\leq h<0$, then $|\theta^{\prime}-\theta|\leq \frac{\pi}{45}$. In summary, with $|h|\leq .065$, the difference between $\theta^{\prime}$ and $\theta$ is smaller than the numerical difference of consecutive angles used in Radon transform computation in our algorithm. Therefore, using equation~\eqref{ver_shear_gam}, we have
\begin{align}
    \label{ver_shear_gam2}
    \hat{s}_{g_2}(t,\theta)=\gamma\hat{s}(t,\theta^\prime)\approx\gamma\hat{s}(t,\theta).
\end{align}

Next we show an approximation for $\gamma$. Apply Taylor's formula for $\sqrt{x}$ around $x = 1$ to $\gamma$, we have that
\begin{align}\label{tayloralpha}
	\gamma &= \sqrt{1+v^2\sin^2\theta+v\sin(2\theta)} \nonumber\\
	&= 1+ \frac{1}{2}\big(v^2\sin^2\theta+v\sin(2\theta)\big)-\frac{\big(v^2\sin^2\theta+v\sin(2\theta)\big)^2}{8\xi^{3/2}},
\end{align} 
where $\xi \in [1, 1+v^2\sin^2\theta+v\sin(2\theta)]$. Ignoring higher order terms we have that for $v\geq 0$
\begin{equation}
	\gamma = 1+ \frac{1}{2}\big(v\sin(2\theta)+v^2\sin^2\theta)\big) + \bigo(v^2),
\end{equation}
with  
\begin{equation}
	|\gamma - 1| \leq\frac{1}{2}(v+v^2)+\frac{1}{8}(v+v^2)^2.
\end{equation}
Hence $\hat{s}_{g_2}(t,\theta) \approx \gamma \hat s(t,\theta) = \hat s(t,\theta)+ \frac{1}{2}(v\sin(2\theta)+v^2\sin^2\theta)\hat s(t,\theta)+\bigo(v^2)$. 
In summary, to model small vertical shearing of $s$ as in \eqref{eqn:eq2} with $v\approx 0$, we add the following additional spanning set $\hat E = \{\big(v^2\sin^2\theta+v\sin(2\theta)\big) \hat s\}$ (for small $v$) as enrichment to the training subspace in the transform space.


\end{document}